\documentclass{article}
\pdfoutput=1
\usepackage{amsmath,amsfonts,bm}

\def\eqref#1{equation~\ref{#1}}
\def\floor#1{\lfloor #1 \rfloor}
\def\1{\bm{1}}

\DeclareMathAlphabet{\mathsfit}{\encodingdefault}{\sfdefault}{m}{sl}
\SetMathAlphabet{\mathsfit}{bold}{\encodingdefault}{\sfdefault}{bx}{n}

\usepackage{arxiv}

\usepackage[T1]{fontenc}    % use 8-bit T1 fonts
\usepackage[pagebackref=true,breaklinks=true,letterpaper=true,colorlinks,bookmarks=false]{hyperref}
\usepackage{url}            % simple URL typesetting
\usepackage{booktabs}       % professional-quality tables
\usepackage{amsfonts}       % blackboard math symbols
\usepackage{nicefrac}       % compact symbols for 1/2, etc.
\usepackage{microtype}      % microtypography
\usepackage{graphicx}
\usepackage{times}
\usepackage{epsfig}
\usepackage{amsmath}
\usepackage{amssymb}
\usepackage{booktabs,siunitx}
\usepackage{xcolor}
\usepackage{comment}
\usepackage{scrextend}
\usepackage{chngcntr}
\usepackage{float}
\usepackage{url}
\usepackage{subfig}
\usepackage{lipsum}
\usepackage{multirow}
\usepackage{algorithm}
\usepackage{algpseudocode}

\definecolor{my_color}{rgb}{0,0,0}
\definecolor{my_color_1}{rgb}{0,0,0}

\begin{document}

\title{Dithered backprop: A sparse and quantized backpropagation algorithm for more efficient deep neural network training}

\newcommand{\addr}{Department of Video Coding \& Analytics, Fraunhofer Heinrich Hertz Institute, Berlin, Germany}

\newcommand{\mails}{\footnotesize{\texttt{\{simon.wiedemann, temesgen.mehari, kevin.kepp, wojciech.samek\} @hhi.fraunhofer.de}}}

\author{
 Simon Wiedemann \\
  \And
 Temesgen Mehari \\
  \And 
  Kevin Kepp \\
  \And
 Wojciech Samek 
}
\maketitle
%%%%%%%%% ABSTRACT
\begin{abstract}
   Deep Neural Networks are successful but highly computationally expensive learning systems. One of the main sources of time and energy drains is the well known backpropagation (backprop) algorithm, which roughly accounts for 2/3 of the computational complexity of training. In this work we propose a method for reducing the computational cost of backprop, which we named dithered backprop. It consists in applying a stochastic quantization scheme to intermediate results of the method. The particular quantisation scheme, called non-subtractive dither (NSD), induces sparsity which can be exploited by computing efficient sparse matrix multiplications. Experiments on popular image classification tasks show that it induces 92\% sparsity on average across a wide set of models at no or negligible accuracy drop in comparison to state-of-the-art approaches, thus significantly reducing the computational complexity of the backward pass. Moreover, we show that our method is fully compatible to state-of-the-art training methods that reduce the bit-precision of training down to 8-bits, as such being able to further reduce the computational requirements. Finally we discuss and show potential benefits of applying dithered backprop in a distributed training setting, where both communication as well as compute efficiency may increase simultaneously with the number of participant nodes.
\end{abstract}
\keywords{Efficient Deep Learning \and Quantisation \and Dither signals \and Distributed Learning}

%%%%%%%%% BODY TEXT %%%%%%%%%
% Introduction.
\section{Introduction}
Deep neural networks (DNNs) are powerful machine learning systems for recognizing patterns in large amounts of data. They became very popular through recent successes in computer vision, language understanding and other areas of computer science \cite{goodfellow2016deep}. However, DNNs need to undergo a highly computationally expensive training procedure in order to extract meaningful representations from large amounts of data. For instance, \cite{DNN_co2} showed that the training process of state-of-the-art neural network architectures can produce 284 tons of carbon dioxide, nearly five times the lifetime emissions of an average car.   Therefore, in order to mitigate the impact of training and/or allow for models to be trained on resource-constrained devices, more efficient algorithms have to be designed. 

The backpropagation (backprop) algorithm \cite{rumelhart1986learning} is most often applied when gradient-based optimization techniques are selected for training DNNs. However, it involves the computation of many dot products between large tensors, therefore playing a major role in the computational cost of the training procedure. Techniques such quantization and/or sparsity can be employed in order to reduce the complexity of the dot products, however, when applied in a na\"ive manner they may induce biased, non-linear errors which can have catastrophic effects for the convergence of the overall training algorithm.

In this work we aim to minimize the computational complexity of the backprop algorithm by carefully studying the error induced by quantization. Concretely, we propose to apply a particular type of stochastic quantization technique to the gradients of the preactivation values, known as non-subtractive dithering (NSD) \cite{schuchman_dither_1964}. NSD does not only reduce the precision of the preactivation values, but it also induces sparsity. As such, we attain sparse tensors with low precision non-zero values, properties that can be leveraged on in order to lower the computational cost of the dot products they are involved in. Our contributions can be summarized as follows:
\begin{itemize}
\item We reduce the computational complexity of the most expensive components of the backprop algorithm by applying stochastic quantization techniques to the gradients of the preactivation values, inducing sparsity + low-precision non-zero values.
\item We show on extensive experiments that we can reach a significant amount of sparsity across a wide set of neural network models while maintaining the non-zero values below/equal to 8-bit precision, without affecting their final accuracy of their model neither the convergence speed. 
\item Finally, we discuss the positive properties that emerge when applying dithered backprop in a distributed setting. Concretely, we show that we can reduce the computational cost for training at each node by increasing the number of participant nodes.
\end{itemize}

%% Related Works.
\section{Related Work}
A lot of research is dedicated to improve the performance at inference time \cite{DLC_survey2, Efficient_DNN_processing, ECT_DNN}. However, less research has focused on designing more efficient training algorithms, in particular a more efficient backward pass. In the following we discuss some of the proposed approaches.

\textbf{Precision Quantization.} Most of preceding work on efficient neural network training uses \textit{Precision Quantization}. In the context of deep learning that means to transform activation, weight and gradient values to representations of lower precision than the regular single-point floating point standard. It has been shown that this can significantly reduce the time and space complexity of deep learning models \cite{courbariaux2014training, courbariaux_binarized_2016, courbariaux_binaryconnect:_2015, micikevicius_mixed_2017, zhou_dorefa-net:_2016, hubara_quantized_2016, NIPS2018_7761}.\\
\cite{courbariaux2014training} were among the first to show that it is feasible to quantize parts of state-of-the-art models without or just with negligible loss of accuracy using 10-bit multiplications. Subsequently, more people followed the example and quantized successfully whole models to 16-bit representations \cite{micikevicius_mixed_2017, gupta_deep_2015}. Later, even ternary and binary weight quantizations were applied, while keeping the gradients and errors in the backward pass in full precision \cite{courbariaux_binaryconnect:_2015, wen_terngrad:_2017}. However, these approaches sacrifice accuracy over the baseline networks. \cite{NIPS2018_7761} accomplished to quantize weights, activations and all gradient calculations, except for the weight updates, to 8-bit. A 16-bit copy of the backpropagated gradient is saved to compute a full-precision weight update. They argue that the extra time required for this matrix multiplication is comparably small to the time required to backpropagate the error gradient and that in most layers these calculations can be made in parallel.

\textbf{Efficient Approximations.}
Other work investigated the possible speed up gaining from efficient approximations of matrix multiplications in the backward pass. \cite{adelman_faster_2018} reduces the complexity of the matrix multiplication by approximations through a form of column-row sampling. Using an efficient sampling heuristic, this approach achieves up to 80\% reduced computation but the authors provide no analysis of the induced noise variance contained the weight gradients and its impact on the generalization performance. The meProp algorithm \cite{sun_meprop:_2017} sparsifies the pre-activation gradients by selecting the $k$ elements with the largest magnitude. They leverage sparse matrix multiplications for a more efficient backward pass. However, since this quantization function is deterministic and operates on vectors, it results in \textit{biased} estimates of the weight updates which can harm the convergence speed as well as generalization performance of the trained model.

In contrast, we show how \textit{dither functions} can be used to calculate \textit{unbiased} weight updates efficiently, due to their sparsity-inducing property when applied to gradient values. Furthermore, we show how the approach can be combined with state-of-the-art precision quantization methods in order to boost the computational efficiency of the algorithm. 
%% Method.
%\input{sections/preliminaries.tex}
\section{Dithered backpropagation}
For fully-connected layers the operations that need to be performed per layer during one training iteration are the following (note that these equations are analogous for convolutional layers):
\begin{align}
\intertext{\textbf{Forward pass}} 
z^l & = W^l\cdot a^{l-1} +b^l
\label{eq:z}
\\
a^l & = f(z^l) \nonumber
\\
\intertext{\textbf{Backward pass}} 
\delta_z^l & = \delta_a^l \odot f'(z^l)  \nonumber
\\
\delta_a^{l-1} & = (W^l)^T \cdot \delta_z^l
\label{eq:da}
\\
\delta_W^l & = \delta_z^l \cdot (a^{l-1})^T
\label{eq:dw}
\end{align}
with $W$, $b$, $z$ and $a$ being the weight tensor, bias, preactivation and activation values respectively. Naturally, $\delta_W$, $\delta_b$, $\delta_z$ and $\delta_a$ denote the error or gradients of the respective quantities. With $f$ we denote the non-linear function whereas with $f'$ its derivative. $l$ is an index referring to a particular layer and $T$ denotes the transpose operation. Finally, the symbols $\cdot$ and $\odot$ denote the dot and Hadamard product respectively. 

As one can see, there are three major matrix multiplications involved at each layer during one training iteration, namely, one in the forward pass (\eqref{eq:z}) and two in the backward pass (\eqref{eq:da} and \eqref{eq:dw}). Since up to 90\% of the computing time is spent on performing these dot product operations \cite{sun_meprop:_2017}, in this work we focus on reducing their computational cost. In particular, notice how the preactivation gradients $\delta_z^l$ are present in both matrix multiplications in the backward pass. Hence, in order to save operations, we apply quantization functions that compresses these gradients. 

\subsection{Non-subtractive dithered quantization (NSD)}
For reasons that will become more apparent in the next section, in this work we propose to apply the following quantization function:
\begin{equation}
\begin{split}
\widetilde{x}	&= Q_{\Delta}(x + \nu)\\
&= \Delta \floor{\frac{x + \nu}{\Delta} + \frac{1}{2}}
\end{split}
\label{eq:nsd}
\end{equation}
with $\Delta$ being the quantization step size and $x\in \mathbb{R}$ some input value.  $\nu \sim U(-\frac{\Delta}{2}, \frac{\Delta}{2})$ is a random number sampled from the uniform distribution between the open interval $(-\frac{\Delta}{2}, \frac{\Delta}{2})$. The quantization function in \eqref{eq:nsd} is sometimes referred as \textit{non-subtractive dither} (NSD)  \cite{schuchman_dither_1964}  in the source coding literature, with $\nu$ being a stochastic \textit{dither signal} that is added to the input before quantization. The main motivation for adding a dither signal before quantization is to decouple the properties of the quantization error $\epsilon = Q_{\Delta}(x+\nu) - x$ from the input signal $x$. For instance, it is known that the quantization error of NSD is unbiased and has bounded variance
\begin{align}
\mathbb{E}[\epsilon] & = 0
\label{eq:unbiased}
\\
\mathbb{E}[\epsilon^2] & < \frac{\Delta^2}{4}
\label{eq:bounded variance}
\end{align}

\subsection{Effects of applying NSD to the gradients}
\begin{figure}[t]
	\centering
	\includegraphics[width=0.6\columnwidth]{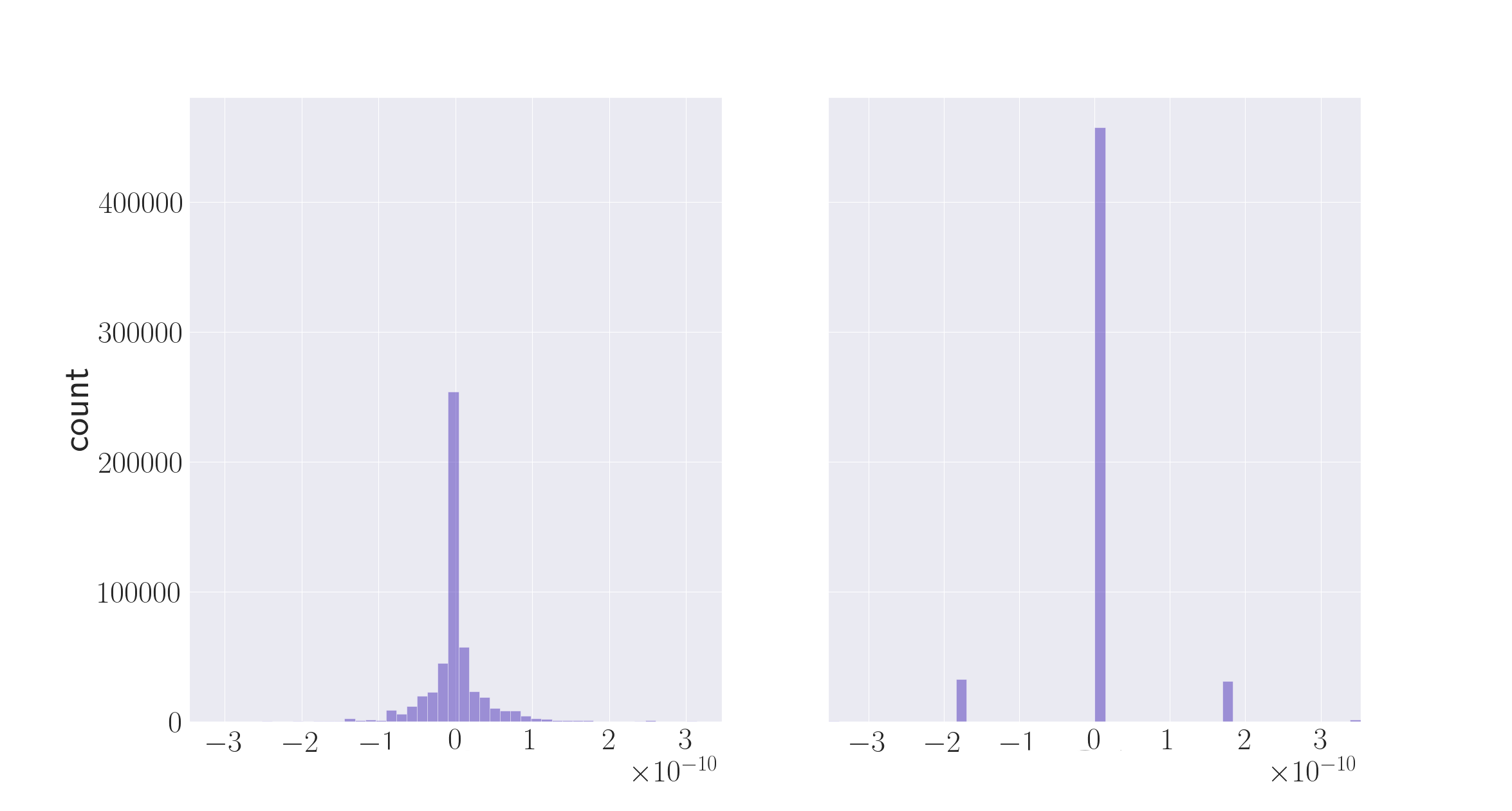}
	\caption{Distribution of preactivation gradient values before $\delta_z$ (left) and after $\widetilde{\delta_z}$ (right) NSD quantization. The gradients have become more sparse (higher count of 0 values) and the non-zero values can be represented with low bitwidths (low number of non-zero ``buckets''). For instance, this example only 1 bit is required to represented all non-zero values.}
	\label{fig:qdz}
\end{figure}

Hence, at each layer $l$, we now apply NSD to the gradients of the preacitvation values $\delta_z^l$ before computing the respective dot products. For large enough stepsizes $\Delta$, NSD will induce sparsity (many zero values) as well as non-zero values with low bitwidth representation (see figure \ref{fig:qdz}). 

\begin{figure}[t]
	\centering
	\includegraphics[width=0.6\columnwidth, height=40mm]{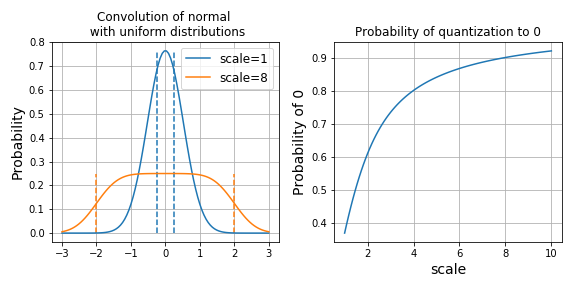}
	\caption{(left) Shape of the probability distribution resulting from the convolution of a gaussian with a uniform distribution, where the uniform distribution samples values between a range $(-\frac{\Delta}{2}, \frac{\Delta}{2})$. The shape depends on the stepsize $\Delta$ of the uniform distribution, which is chosen to be $\Delta = s\sigma$ with $\sigma$ being the standard deviation of the gaussian distribution and $s\in \mathbb{N}$ a scaling factor. The dashed lines indicate the region of values between $(-\frac{\Delta}{2}, \frac{\Delta}{2})$. (right) the probability of a 0 value appearing after quantization at different scale factors. It is calculated by integrating the area between the dashed lines on the left plot. From both plots one can see that sparsity increases with the scaling factor $s$.}
	\label{fig:conv}
\end{figure}

To make this effect more clear, consider the convolution $f(t) = (G_{\sigma}*U_{\Delta})(t)$ between $G_{\sigma}$ a gaussian distribution with mean 0 and standard deviation $\sigma$ and $U_{\Delta}$ a uniform distribution, sampling values in the range $(-\frac{\Delta}{2}, \frac{\Delta}{2})$. The induced average sparsity is given by the probability of $f$ sampling a value in the same interval, thus
\begin{equation*}
P(t=0) = \int_{-\frac{\Delta}{2}}^{\frac{\Delta}{2}}f(t)
\end{equation*}
As figure \ref{fig:conv} shows, the probability of 0 increases with the stepsize value. Naturally, the same applies for the maximal bit-width of the non-zero values since the probability of a high number appearing after quantization decreases as the stepsize increases.

%%%%%%%%%%%%%%%%%%%%%%%%%%%%%%%%
\iffalse
\begin{align}
f(t) & = (G_{\sigma}*U_{\Delta})(t) \nonumber
\\ 
& =  \frac{1}{2\Delta}\left[ \text{erf}\left(\frac{1}{\sigma\sqrt{2}}(t + \Delta )\right) -\text{erf}\left(\frac{1}{\sigma\sqrt{2}}(t-\Delta )\right) \right]
\label{eq:conv}
\end{align}

\begin{align*}
P(t=0) & = \int_{-\frac{\Delta}{2}}^{\frac{\Delta}{2}}f(t) \\
	  & = \text{erf}(\frac{3\Delta}{2}) - \text{erf}(\frac{\Delta}{2})
\end{align*}
\fi
%%%%%%%%%%%%%%%%%%%%%%%%%%%%%%%%

We can then exploit this sparsity to omit operations when computing the dot product between tensors. The altered equations for the backward pass at each layer are then given by:
\begin{align}
\widetilde{\delta_z^l} & = Q_{\Delta^l}(\delta_z^l + \nu^l)
\label{eq:qdz}
\\
\delta_a^{l-1} & = (W^l)^T \cdot \widetilde{\delta_z^l}
\label{eq:qda}
\\
\delta^l_W & = \widetilde{\delta_z^l} \cdot (a^{l-1})^T
\label{eq:qdw}
\end{align}
with $\nu^l \sim U(-\frac{\Delta^l}{2}, \frac{\Delta^l}{2})$ and $\widetilde{\delta_z^l}$ being the matrix of quantized pre-activation gradients.

Given the above analysis we propose to choose the stepsize at each layer as to be a multiple of the standard deviation, that is,  $\Delta^l = s\sigma^l$ $\forall l$, with $\sigma^l$  being the standard deviation of the preactivation gradients and $s\in \mathbb{N}$. $s$ is thus a global scaling factor that controls the trade-off between compute complexity and learning performance. We named our proposed modification of the backprop method \textbf{\textit{dithered backprop}}. Algorithm \ref{alg:dithered backprop} shows a pseudocode of the quantization procedure of the preactivation gradients. After quantization, the backward pass as well as the weight update steps remain identical as in the usual algorithm.

\begin{algorithm}[t]
    \caption{Dithered backprop quantization}
     \begin{algorithmic}[1] % The number tells where the line numbering should start
        \Procedure{NSD}{$\delta_z^l$, $s$} \Comment{Quantizes preactivation gradients $\delta_z^l$ of layer $l$}
            \State $\sigma^l \gets \text{std}(\delta_z^l)$ \Comment{Computes standard deviation}
            \State $\Delta^l \gets s\sigma^l$
            \State $\widetilde{\delta_z^l} \gets Q_{\Delta}(\delta_z^l)$ \Comment{As in \eqref{eq:nsd}}
            \State \textbf{return} $\widetilde{\delta_z^l}$
        \EndProcedure
    \end{algorithmic} 
    \label{alg:dithered backprop}
\end{algorithm}

\subsection{Error statistics and convergence of the method}
Due to applying NSD to all $\delta_z^l$, dithered backprop attains perturbed estimates of the weight updates 
\begin{equation*}
\widetilde{\delta_W^l} = \delta_W^l + \epsilon_W^l
\end{equation*}  
with $\epsilon_W^l$ being the perturbation error. Hence, this begs the question: how does this error influence the convergence of the training method?

From \cite{Bottou98onlinelearning} we know that under mild assumptions regarding the loss function, if a stochastic operator is added to a training algorithm that already converges and generates unbiased estimates of the weight updates with bounded variance, then the respective training algorithm converges as well. Thus, we only need to show that the error of the weight updates is unbiased and has bounded variance, that is
\begin{align}
\mathbb{E}[\epsilon_W^l] & = 0 \; \forall l
\label{eq:dw_unbiased}
\\
\mathbb{E}[(\epsilon_W^l)^2] & < C \; \forall l
\label{eq:dw_bounded variance}
\end{align}
Although in this work we do not provide a rigorous proof (mainly due to space constraints), it is relatively easy  to see that \eqref{eq:dw_unbiased} and \eqref{eq:dw_bounded variance} are satisfied by modelling the quantization error of the preactivation gradients also as additive noise $\widetilde{\delta_z^l} = \delta_z^l + \epsilon_z^l$, and taking into consideration that the error $\epsilon_z^l$ satisfies \eqref{eq:unbiased} and \eqref{eq:bounded variance}.

%%%%%%%%%%%%%%%%%%%%%%%%%%%%%%%
\iffalse
\textbf{Theorem}. \textit{When a quantization scheme as described in \eqref{eq:nsd} is applied to each element of the gradients of the preactivation values as described in \eqref{eq:qdz}, and if the weight updates are accordingly computed as described in \eqref{eq:qda} and \eqref{eq:qdw}, then the error $\epsilon_W^l = \widetilde{\delta_W^l} - \delta_W^l$ of each weight update induced by the quantization at each layer $l$ satsifies the following properties:}
\begin{align}
\intertext{\textbf{Unbiased}} 
\mathbb{E}[\epsilon_W^l] & = 0 \quad \forall l
\end{align}
\textit{and}
\begin{align}
\intertext{\textbf{Bounded variance}} 
\begin{split}
&\qquad \mathbb{E}[(\delta dW_{ij}^l)^2] \le \\
&\frac{1}{B^2} \sum_{k=1}^{B} 
f'(a_{ik}^l)^2\mathbb{E}[(\delta da_{ik}^l)^2] (a_{jk}^{l-1})^2
+ \frac{\Delta^2}{4B^2} ||a_{j*}^{l-1}||_2  ,
\end{split}
\label{eq:cor_dW}
\end{align}
\textit{with}
\begin{equation}
\begin{split}
& \qquad \mathbb{E}[(\delta da_{jk}^{l-1})^2] \le \\
& \sum_{k=1}^{n_l}(W_{ij}^l)^2 f'(a_{ik}^l)^2\mathbb{E}[(\delta da_{ik}^l)^2] 
+ \frac{\Delta^2}{4} ||W_{j*}^l||_2.
\end{split}
\label{eq:cor_da}
\end{equation}
The variance bound in equations \ref{eq:cor_dW} and \ref{eq:cor_da} follows from the fact that uniform quantization with uniform dither induces quantization noise with bounded variance. \textcolor{red}{(TODO) The formal proof can be found in the appendix}.
\fi
%%%%%%%%%%%%%%%%%%%%%%%%%%%%%%%%

\subsection{Computational complexity}
\paragraph{Theoretical analysis}$\;$\\
When dithered backprop is used for training, some additional computational overhead comes form applying NSD to the gradients of the preactivation values. However, we argue that this cost is asymptotically negligible compared to the cost of performing the subsequent dot products. In the following we will highlight the rationale for the case of fully-connected layers, however, we stress that it also applies analogously to convolutional layers. 

Let $G$ be a $(k\times n)$-dimensional matrix whose elements are the gradient of the preactivation values of a particular layer. As can be seen from \eqref{eq:nsd} and algorithm \ref{alg:dithered backprop}, applying NSD to $G$ requires: for each element,
\begin{enumerate}
\item calculate the standard deviation of the preactivation gradients. This requires 1 multiplication + 1 addition per element.
\item sampling from the uniform distribution between the interval $(-\frac{\Delta}{2}, \frac{\Delta}{2})$. This requires about 2 multiplications + 2 additions + 1 modulo operation.
\item Quantizing the value, which requires 1 addition + 1 multiplication + truncation of decimal bits
\end{enumerate}
Overall, the cost can be approximately reduced to about 9 arithmetic operations per element. Thus, the computational complexity of applying NSD is of order $\mathcal{O}(kn)$. If we now include the cost of performing the subsequent sparse matrix-matrix dot product, then the complexity becomes of order $\mathcal{O}(kn + p_{nz}mkn)$, with $p_{nz}$ being the empirical probability of non-zero values in $G$.

In contrast, the computational complexity of a matrix multiplication of the form $W\cdot G$ with $W$ being, for instance, an arbitrary $(m\times k)$-dimensional weight matrix, is of order $\mathcal{O}(mkn)$. If we now measure the effective asymptotic savings between the dithered dot product vs the dense dot product algorithm by taking the ratio of both quantities we get
\begin{equation}
\text{comp. savings} = \mathcal{O}\left(\frac{1}{m} + p_{nz}\right) \xrightarrow{m \gg 1} \mathcal{O}(p_{nz}) 
\label{eq:gains}
\end{equation}
The above \eqref{eq:gains} states that the asymptotic computational savings depend inversely on the amount of rows $m$ of the output matrix, as well as on the sparsity attained after applying NSD. Since the number of output rows $m$ are most often much bigger than one, the computational savings are dominated solely by the sparsities achieved. Later in the experimental section we show that NSD is able to induce high sparsity ratios (between 75\% - 99\%) during the entire training procedure, thus in principle being able to achieve significant savings. 

\paragraph{Practical savings}$\;$\\
Unfortunately, the above analysis does not translate directly to real-world speedups/energy savings mainly due to the challenges that unstructured sparsity imposes on the hardware level. Nevertheless, it is worth to mention that in recent years there has been significant progress in this field, showing promising results in narrowing the gap between the theory and practice. On a software level, \cite{elsen2019fast} have shown that they can already attain up to x2.4 speedups for DNNs with 80\%-90\% sparsity, by optimizing the sparse dot products so that it becomes more amenable to the underlying hardware. On the other hand, many hardware accelerators have been proposed \cite{EIE, EmbeddedDNN_book, eyerissv2, scnn} that are able to successfully exploit structured and unstructured sparsity, sometimes achieving orders of magnitude more compute efficiency. In particular, \cite{scnn} attained about x1.5-x8 speedups and x1.5-x6 energy gains at sparsity ratios between 75\%-95\%, ratios that are typically induced by dithered backprop (see experiments section). Finally,  \cite{DitherNN} proposed an accelerator that includes an efficient implementation of dithered quantization in order to execute DNNs with lower bit-precision. Hence, this progress motivates the study of methods akin to dithered backprop, since it seems likely to expect similar gains when such algorithms are implemented in an efficient manner on a software level and run on similarly optimized hardware architectures.

\subsection{Quantizing forward pass}
So far we have only discussed the reduction of the computational cost of the backprop method. Although the backward pass accounts for roughly 2/3 of the computational complexity of the training iteration (see \eqref{eq:z}, \eqref{eq:da}, \eqref{eq:dw}), we are also interested in applying methods that also reduce the computation of the forward pass. Fortunately, some research has already been done in this area.
\\
 \cite{NIPS2018_7761}, e. g., quantizes activation, weight and some gradient values in the backward pass to 8-bits and show that using their method state-of-the-art results can still be achieved. In addition, they introduced Range Batch-Normalization (BN), an approximated batch norm that scales a batch by dividing it by its range. It has significantly higher tolerance to quantization noise and improved computational complexity. 
 \\
Armed with this knowledge, we similarly quantize activation and weight values in the forward pass and apply dithered backprop in the backward pass, leaving also only the weight update in full precision. Therefore, all compuations, except for the weight update, can be calculated with 8-bit computations.

\subsection{Usage in distributed training setting}
\label{subsec: distributed training}
A further interesting area of application of the dithered backprop method is distributed training. In distributed training, an algorithm called \textit{synchronous stochastic gradient descent} (SSGD) is widely used \cite{sattler2020trends}. It differs from single-threaded mini-batch SGD in that the mini-batch of size $m$ is distributed to $N$ total workers that locally compute sub-mini-batch gradients. These gradients are then communicated to a centralized server called \textit{parameter server}  that updates the parameter vector and then eventually sends it back. By increasing the number of training nodes and taking advantage of data parallelism, the total computation time of the forward-backward passes on the same size training data can as such be dramatically reduced.
\\
As mentioned in the above section, dithered backprop induces unbiased noise with bounded variance to the weight updates. Therefore, if dithered backprop is applied to $N$ nodes, then most of the induced noise cancels out on the server side due to the averaging effect. Moreover, the variance of the noise goes down with $1/N$. Thus, dithered backprop promises to reduce the computational cost per node as the number of nodes $N$ grows, since stronger quantization can be applied without affecting the final  performance of the model after training. This may be beneficial for scenarios where a large number of nodes with limited computational resources may participate in the training procedure, e.g., a large number of mobile devices connected through a communication channel with high bandwidth such as 5G.

%% Experiments \& Results.
\section{Experiments}
\begin{table*}[t]	
	\begin{center}
		\resizebox{\columnwidth}{!}{
		\begin{tabular}{c|c|cc|cc||cc|cc} \hline
			\multirow{2}{*}{Model} & \multirow{2}{*}{Dataset} &  \multicolumn{2}{c}{Baseline}  &  \multicolumn{2}{c}{Dithered Backprop} &  \multicolumn{2}{c}{8-bit Training  \cite{NIPS2018_7761}} &  \multicolumn{2}{c}{8bit + dith. backprop}  \\\cline{3-10}
			& & 		acc\% & sparsity\% & acc\% & sparsity\% & acc\% & sparsity\% & acc\% & sparsity\% \\ \hline
			LeNet5      &   MNIST      	& 99.31     &   2.05	&   \textbf{99.35}   &   \textbf{97.52}   &   99.34   &   2.09   	&   99.35   &   97.18   \\
			LeNet300100 &   MNIST      	& 98.45     &   47.48   &   98.40   &   \textbf{94.92}   &   98.43   &   48.61   &   \textbf{98.52}   &   94.85   \\
			
			AlexNet     &   CIFAR10    	& 91.23     &   91.35   &   \textbf{91.26}   &   \textbf{98.95}   &   91.03   &   64.62   &   90.81   &   97.05   \\
			ResNet18    &   CIFAR10		& \textbf{92.67}     &   24.36   &   92.35   &   91.86   &   92.22   &   34.88   &   92.10   &   \textbf{92.10}   \\
			VGG11       &   CIFAR10    	& 92.35     &   8.47   	&   92.17  	&   94.10   &   \textbf{92.44}   &   4.82   	&   92.29   &   \textbf{94.24}   \\

			AlexNet     &   CIFAR100   	& 67.98     &   92.23   &   67.78   &   \textbf{97.35}   &   \textbf{68.37}   &   64.39   &   67.63   &   89.51   \\
			Resnet18    &   CIFAR100    & 69.54     &   18.23   &   69.97   &   87.66   &   \textbf{70.73}   &   13.39   &   69.69   &   \textbf{87.74}   \\
			VGG11       &   CIFAR100   	& 70.58     &   6.70   	&   70.09   &   \textbf{91.79}   &   \textbf{71.29}   &   83.40   &   70.07   &   91.77   \\
			
			Resnet18    &   ImageNet    & \textbf{71.40}     &   6.44   &   71.10   &   \textbf{75.80}   &   71.25   &   3.27   &   71.23   &   75.48   \\
			\hline
			Average    &   -    & 83.72     &   33.03   &  83.61  &   \textbf{92.22}   &   \textbf{83.90}  &   35.50   &   83.52  &   91.10   \\
			Average diff.    &   -    &  0   &   0  &  0.23 &   59.12   &   0  &  0   &   0.40 &   55.61   \\
			
		\end{tabular}
		}
		\caption{Results of experiments, where acc\% means accuracy in \% on test set and sparsity\% the average sparsity of the gradients of the preactivation values in \% over all layers and training iterations. The largest values for are marked in bold.}
		\label{tab:results}
	\end{center}
\end{table*}
\textbf{Datasets.} We conducted our experiments on four different benchmark datasets for image classification, namely MNIST, CIFAR10, CIFAR100 and ImageNet.

\textbf{Training Setting.} For the MNIST Dataset Lenet300100 and Lenet5 were evaluated, while for CIFAR10 and CIFAR100 it was VGG11, AlexNet and ResNet18 and for ImageNet only ResNet18. For the CIFAR datasets, we reduced the capacity of the models to account for the dataset. That is, for AlexNet we reduced the dimensionality of the last two hidden layers to 2048, and for VGG11 to 512. The last layers are adapted to account for the classes, respectively.\\

All Models are trained via stochastical gradient descent with a \texttt{momentum} of 0.9, a \texttt{weight} \texttt{decay} of $5 \times 10^{-4}$, and a \texttt{batch}-\texttt{size} of 256 for ImageNet and 128 for the others. We used a learning rate \texttt{lr} of 0.05 for AlexNet and 0.1 for the rest of the models. For the CIFAR datasets a \texttt{lr}-\texttt{decay} setting of 0.1/100 and 0.1/45 is applied.

\subsection{Accuracy \& Induced Sparsity}
For the listed data sets we conducted experiments for four different methods, according to the training setting described above. Besides the baseline method, which describes training without quantization, we applied dithered backprop as described in the above section, the precision quantization of \cite{NIPS2018_7761} (8-bit training) which applies quantization in the forward and backward pass in order to perform training in 8-bit precision, and the combination of the latter two. Table \ref{tab:results} summaries our findings. 

Firstly, notice how the baseline training method exhibits vastly different sparsities across different models, ranging from 2\% to 92\%. Models trained without batchnorm such as AlexNet exhibit already high sparsity ratios due to the derivative of the ReLU activation function, which is often 0. However, batchnorm layers cancel out this effect and therefore models such as LeNet5 or VGG11 exhibit high density (low sparsity). We see a similar effect on models trained with 8-bit precision. On average, the baseline backprop method was able to induce only 33\% sparsity across the different models, and similarly the 8-bit backprop method only 36\%. 

In contrast, after applying dithered backprop, sparsity becomes very high across all networks, ranging between 76\%-99\%. In particular, notice how dithered backprop is able to significantly increase the sparsity of networks trained with batchnorm layers. For instance, LeNet5 goes from 2.05\% to 97.52\%, a substantial increase of 95.47\%.  On average, \textbf{dithered backprop was able to induce 92\% sparsity} across the models, \textbf{increasing the sparsity ratio by 59\%} from the baseline. We get similar results when applied in combination with the 8-bit training method. Here, dithered backprop increased the sparsity by 56\%, inducing an average sparsity of 91\% across the networks. If we consider the speedups and energy gains reported in \cite{scnn}, these results may potentially translate to x5 speedups and x4.5 energy gains on average if dithered backprop is run on specialized hardware.

\begin{figure*}
	\centering
	\subfloat[]{
		\includegraphics[width=0.5\columnwidth]{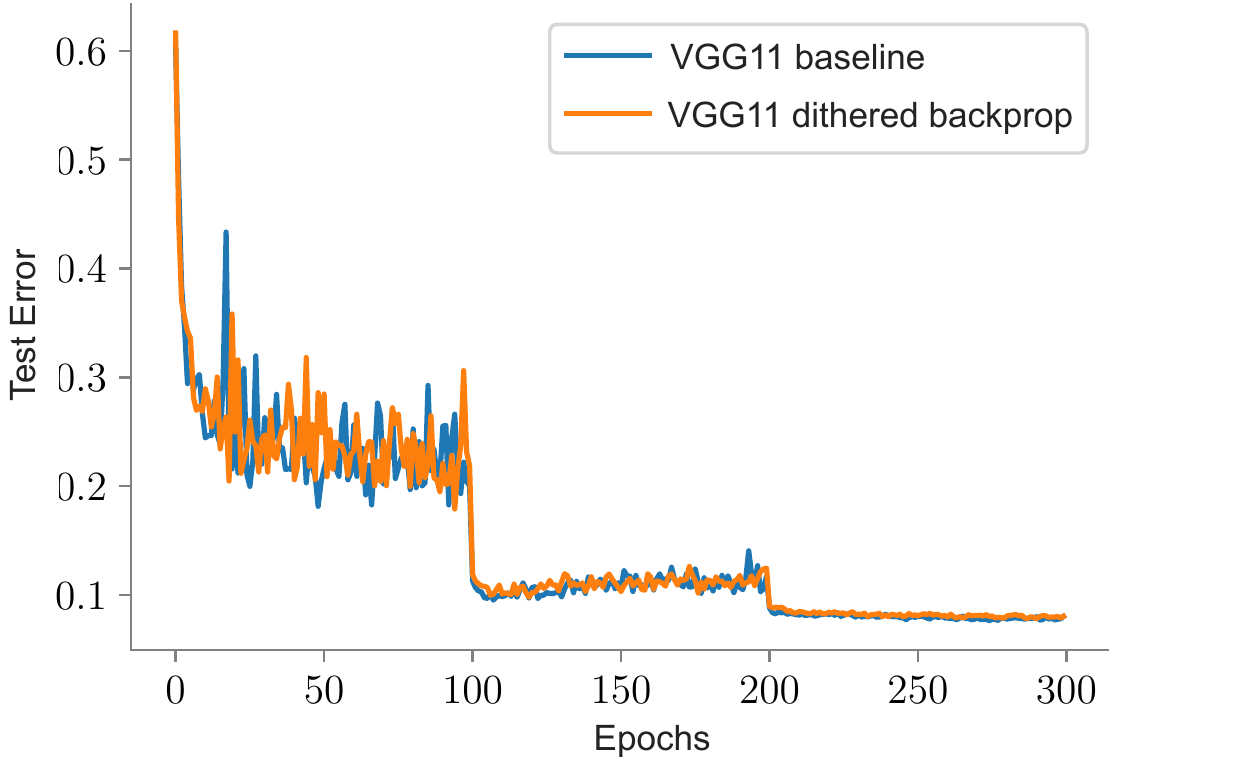}
	}
	\subfloat[]{
		\includegraphics[width=0.5\columnwidth]{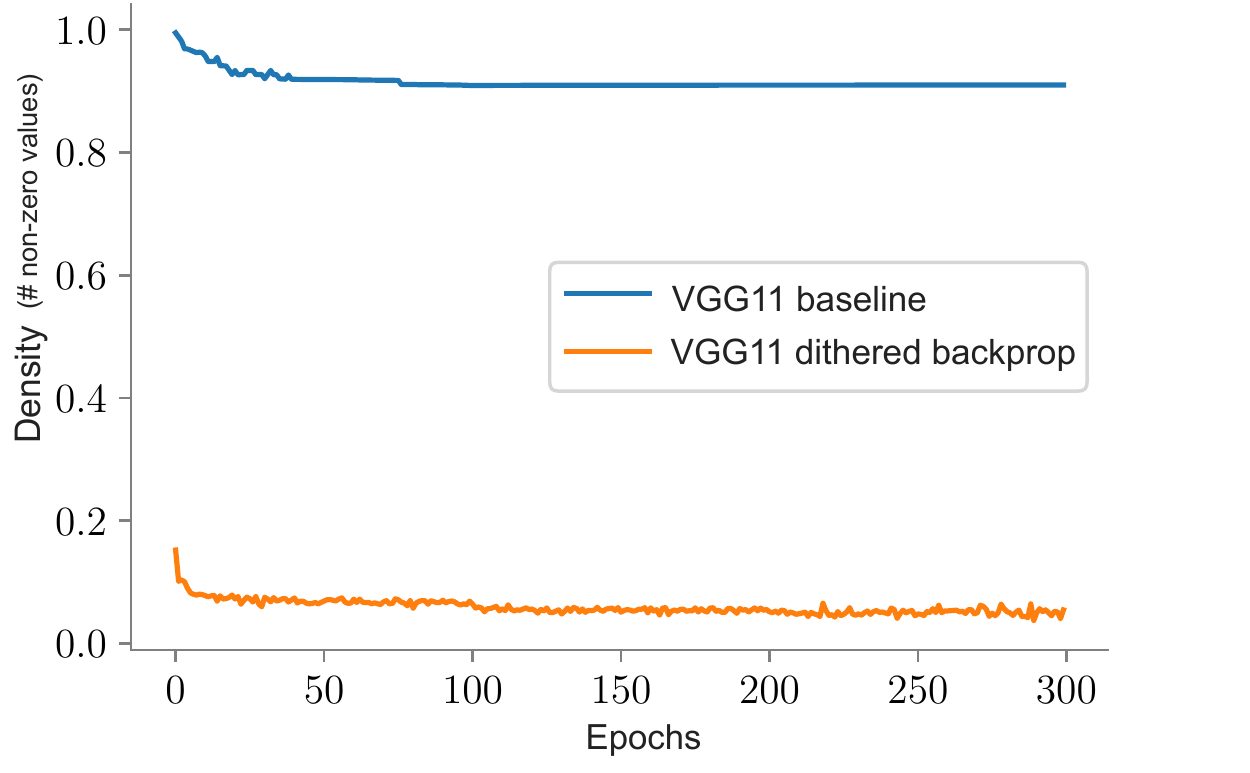}
	}
	\caption{(a) Test error of VGG11 trained on CIFAR10 over the training epochs. (b) Average density (\# non-zero values or 1-sparsity) of the preactivation gradients during training.}
	\label{fig:acc_sparsity}
\end{figure*}

We stress that the accuracies were approximately maintained across the experiments, changing on average only by 0.3\% between the dithered and non-dithered methods. Moreover, the number of training epochs did also not change, showing that \textbf{dithered backprop did not harm the convergence speed}. Figure \ref{fig:acc_sparsity}a shows an example where the test error of AlexNet is plotted over the training epochs. As can be seen, there is no recognizable difference in convergence speed between the baseline model and the dithered model. More examples can be found in the appendix. 

Additionally, we also want to mention that the maximum bitwidth of the non-zero values was consistently below/equal to 8-bits (see figure \ref{fig:DT_sparsity_bits}b) across all experiments. Thus, dithered backprop is fully compatible to training methods that limit the bit-precision training to 8-bits, such as \cite{NIPS2018_7761}.

In Figure \ref{fig:acc_sparsity}b we show the course of the density (\# non-zero values or 1-sparsity) of the preactivation gradient over the entire training period of the VGG11 model. We can see how the density of the gradients is much lower when dithered backprop is applied  across the entire training procedure. Interestingly, we also see that the density decreases at the beginning of training and stays approximately constant afterwards. Coincidentally, this trend correlates weakly with the speed of the learning progress, which can be interpreted as gradients carrying more information. However, it seems that dithered backprop is successful at eliminating redundant, non-useful information for learning since its density is much lower.

\subsection{Comparison to meProp}
 \begin{figure}[t]
	\centering
	\includegraphics[width=0.5\linewidth]{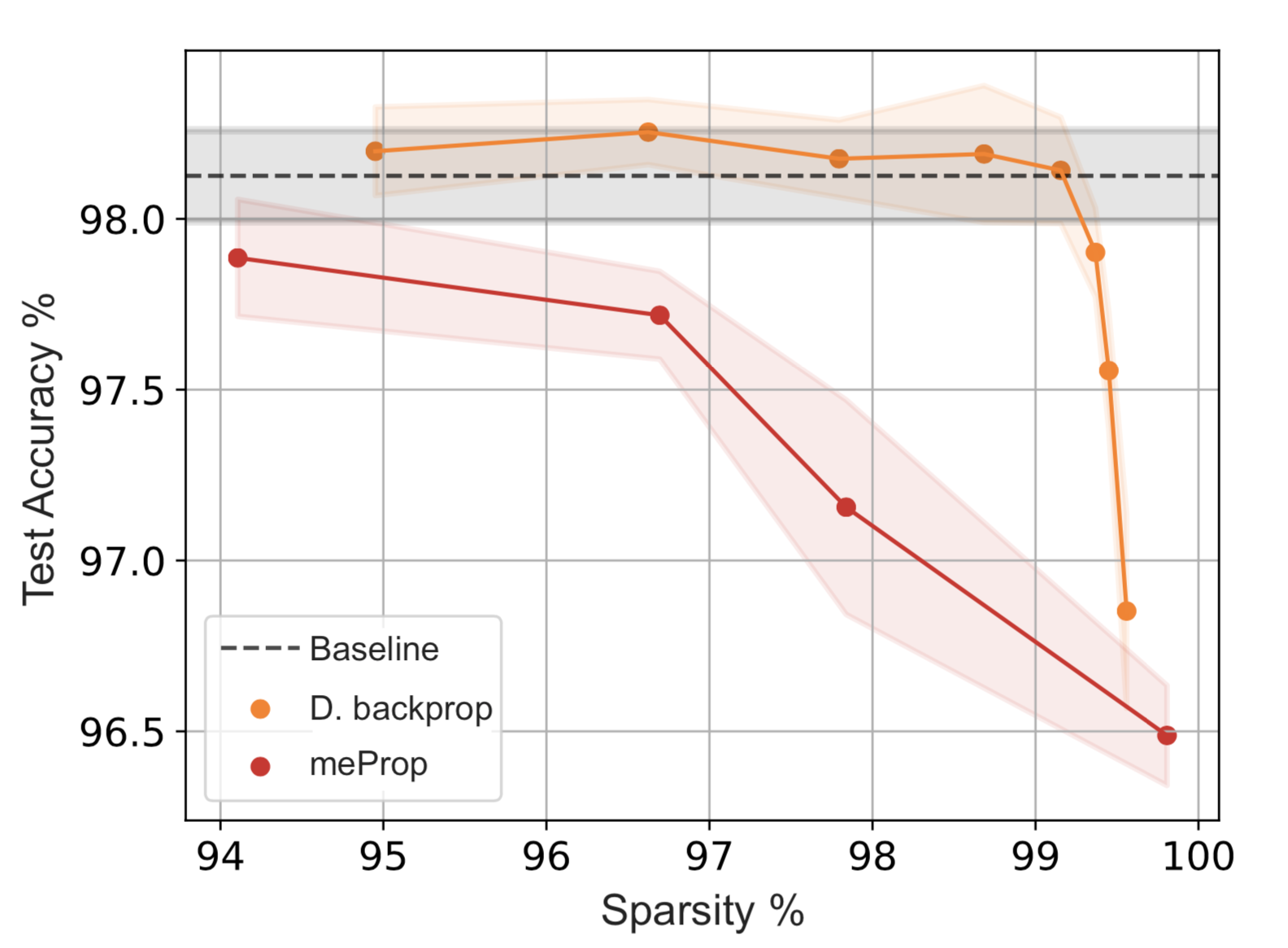}
	\caption{Learning performance at different levels of average sparsity the preactivation gradients of a multilayer perceptron with two hidden layers (500, 500) trained on MNIST, using either regular back propagation (Baseline), dithered backprop (D. backprop) or meProp \cite{sun_meprop:_2017}. Multiple runs with different random seeds were executed for each configuration. Points show mean performance with standard deviation indicated as span.}
	\label{fig:mnist_mepropvsdb}
\end{figure}

We now benchmark dithered backprop against meProp  \cite{sun_meprop:_2017}, arguable the closest related work. To recall, in one of its modes meProp sparsifies the pre-activation gradients by selecting the $k$ elements with the largest magnitude. This induces biased estimates of the weight updates, which we argue affects negatively the learning quality of the network. 

Since meProp was only benchmarked on multilayer perceptrons, we chose a model with two fully-connected layers with hidden dimensions of (500, 500) and trained it on MNIST and CIFAR10 for the experiments. Figure \ref{fig:mnist_mepropvsdb} shows the final test accuracy of the model trained on MNIST at different levels of average sparsity of the preacitvation gradients. On the appendix we show the results for CIFAR10. As one can see, dithered backprop clearly outperforms meProp at all levels of sparsity. Concretely, overall dithered backprop achieved an average test accuracy of $98.14\%$ at a sparsity of $99.15\%$, whereas meProp only achieved $97.89\%$ average test accuracy at a sparsity of $94.11\%$.

\subsection{Distributed training} 
\begin{figure}[t]
	\centering
	\includegraphics[width=0.5\linewidth]{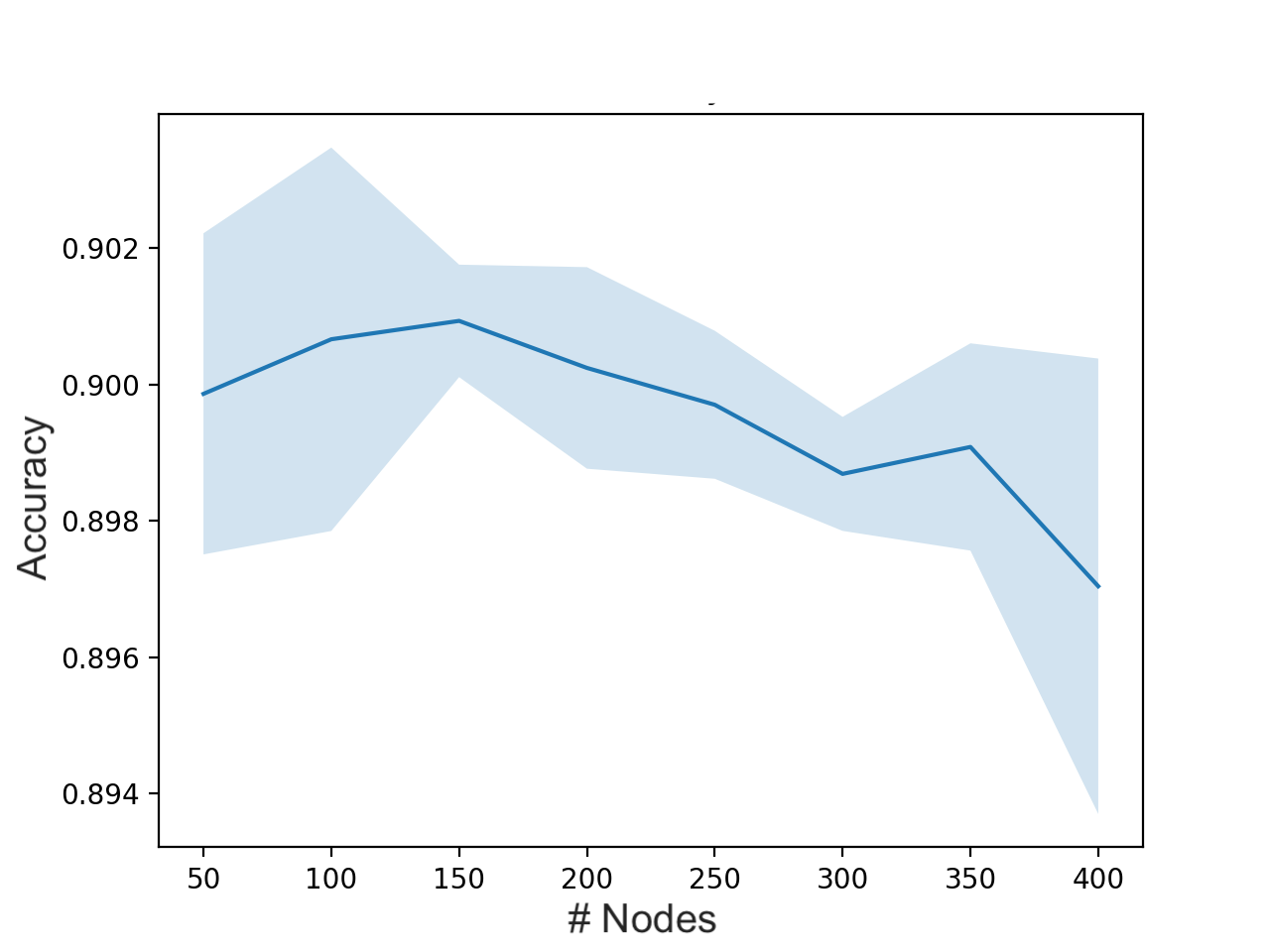}
	\caption{Accuracy of the final model of AlexNet trained on CIFAR10 with dithered backprop in a distributed training setting, at different number of participating nodes configuration. The accuracy stays approximately constant as the number of nodes increases.}
	\label{fig:DT_accuracy}
\end{figure}

\begin{figure*}
	\centering
	\subfloat[]{
		\includegraphics[width=0.5\columnwidth]{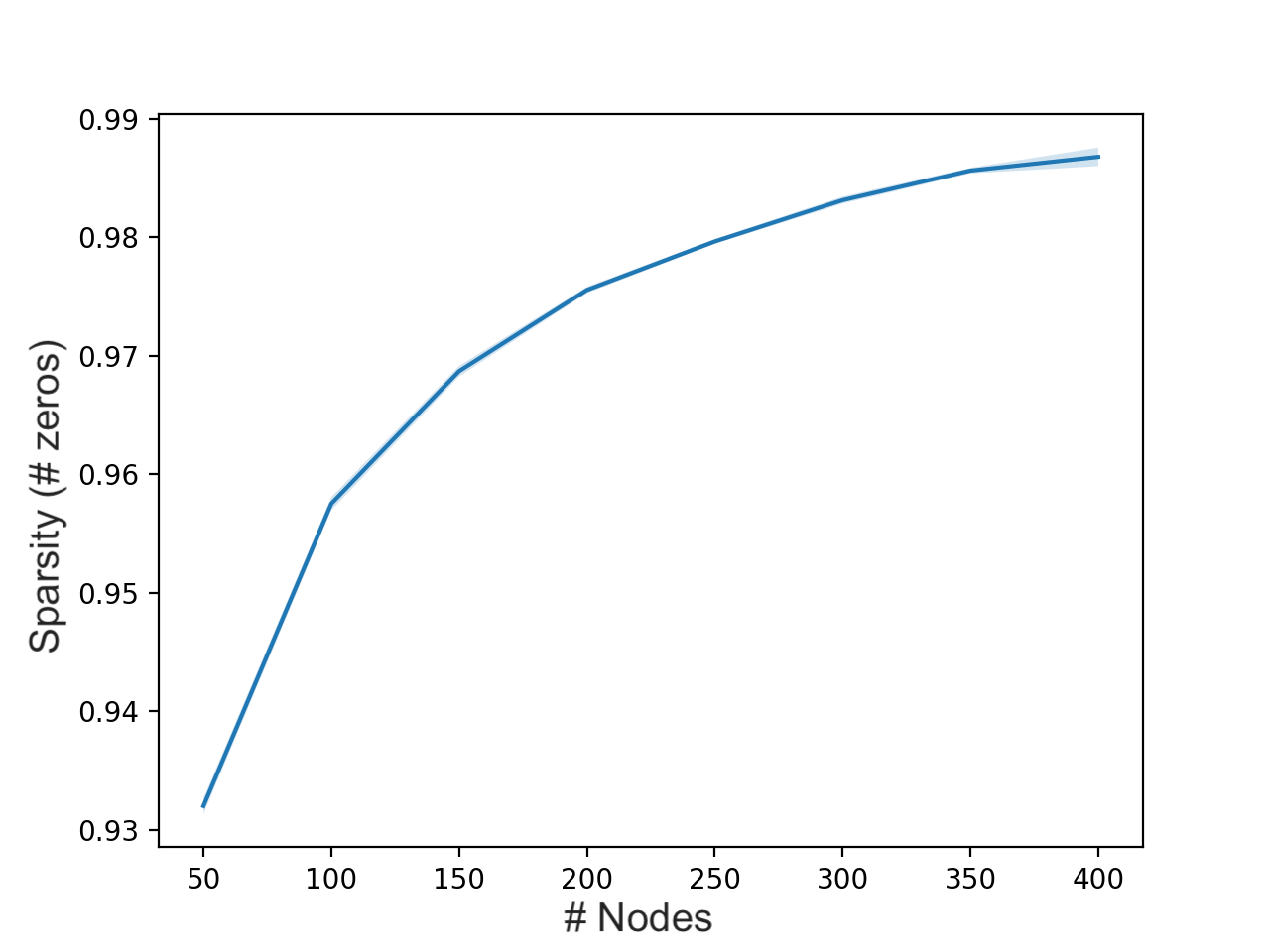}
	}
	\subfloat[]{
		\includegraphics[width=0.5\columnwidth]{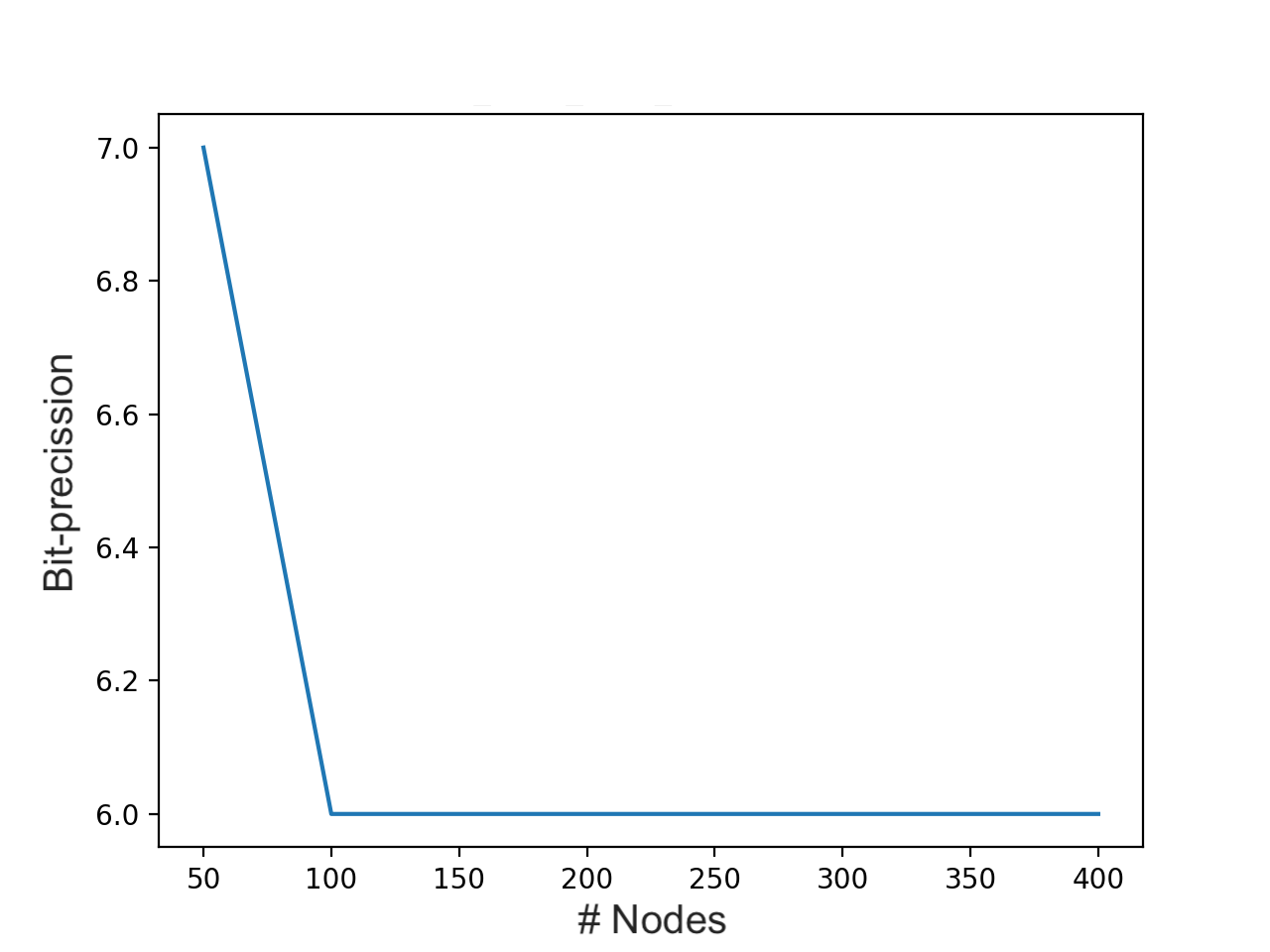}
	}
	\caption{(a) Average sparsity of the preactivation gradients of the fully-connected layers of AlexNet trained on CIFAR10 with dithered backprop in a distributed training setting, at different number of participating nodes configuration. As the number of nodes increases, so does the sparsity at each node and therefore its computational savings for training. (b) Maximal, worst-case bit-precision  of the fully-connected layers of AlexNet trained on CIFAR10 with dithered backprop in a distributed training setting, at different number of participating nodes configuration. As the number of nodes increases, the number of bits necessary to represent the non-zero values decreases, and with it the computational cost for training at each node.}
	\label{fig:DT_sparsity_bits}
\end{figure*}

In the above section we argued that applying dithered backprop in a distributed training scenario may be beneficial. The rationale was that, since the noise induced by dithered backprop on the weight updates is unbiased with bounded variance, then this should cancel out as the number of nodes grows due to the averaging of the parameters on the server. In this section we try to show this effect experimentally.  
\\
To investigate this, we ran several experiments of the same model with different amount of nodes $N$. While increasing $N$, we also increase the scaling factor $s$ of the dither method in order to increase the quantization strength. At each training iteration, each node runs one forward and dithered backward pass of batchsize 1, then sends its parameter gradients to the server where it is subsequently averaged with the gradients of all other nodes. Finally the averaged parameter gradient are broadcasted back to each node, and a new training iteration can subsequently start again. We then measure the final accuracy of the model, average sparsity and worst-case bit-precision at all $N$ configurations. 

Figures \ref{fig:DT_accuracy}, \ref{fig:DT_sparsity_bits} show the respective trends for the fully-connected layers of AlexNet trained on CIFAR10. On the appendix we show the same plots for the convolutional layers as well. Each plot shows the average trend and the standard deviation over 3 different runs of the same experiments. As one can see, we can increase the sparsity and lower the bit-precision as the number of participating nodes $N$ increases, while negligibly affecting the final accuracy of the model. In other words, \textbf{dithered backprop allows to to reduce the computational cost of performing a training iteration at each node as the number of participant nodes increases}. 

As a side note we want to remark that in this context, high sparsities on the preactivation gradients do necessarily translate to communication savings. For batchsizes bigger one the weight updates are with high probability densely populated, so that the full model would have to be communicated at each iteration. Only when the batchsize per node is equal to one (as was in the case of our experimental setup), sparsities on the preactivation gradients directly translate to sparsity on the weight updates and consequently to savings in communication cost.  
%% Conclusions.
\section{Conclusion}
In this work we proposed a method for reducing the computational complexity of the backpropagation (backprop) algorithm. Our method, called \textit{dithered backprop}, is based on applying dithered quantization on the tensor of the preactivation gradients in order to induce sparsity and non-zero values with low bit-precision. It is also simple in that it has only one global hyperparameter which controls the trade-off between computational complexity and learning performance of the model.

Extensive experimental results show that dithered backprop is able to attain high sparsity ratios, between 75\%-99\% across a wide set of neural network models, boosting the sparsity by 59\% on average from the original backprop method. In addition, we showed that dithered backprop maintains the bit-precision of the non-zero values to less/equal 8-bits during the entire training process, thus being fully compatible with methods that limit the training to 8-bit precision only. However, in its current form, dithered backprop induces unstructured sparsity which is not amenable to conventional hardware such as CPUs or GPUs. In future work we will consider modifications that translate directly to speedups and energy gains without having to rely on specialized hardware. Moreover, we will also consider applying efficient compression algorithms to the gradients in order to reduce memory complexity of training as well \cite{8725933, 8970294}.

We also showed that beneficial properties emerge when dithered backprop is applied in the context of distributed training. For instance, we showed experimentally that as the number of participating nodes increases, so does the computational savings per node as well. This effect can be advantageous when a large number of nodes with resource-constrained computational engines participate in the training procedure, such as mobile phones. A further interesting future work direction is to apply dithered backprop jointly with methods that drastically reduce the communication cost \cite{8889996, 8852172}, with the goal of minimizing both the communication as well as computation cost of the distributed training system.

%%%%%%%%% Acknowledgement %%%%%%%%%
%\section*{Acknowledgement}
%This work was funded by the German Ministry for Education and Research as BIFOLD 
%- Berlin Institute for the Foundations of Learning and Data (ref. 01IS18025A and ref 01IS18037I).
%%%%%%%%% Bibliography %%%%%%%%%
{\small
\bibliographystyle{unsrt}
\bibliography{egbib}
}
\newpage
\appendix
\counterwithin{figure}{section}
\counterwithin{table}{section}
\section*{Appendix}
\subsection*{More experimental results}
In this section of the appendix we show further experimental results.

\subsubsection*{Convergence of dithered backprop}
\begin{figure}
\centering
	\includegraphics[width=0.8\columnwidth]{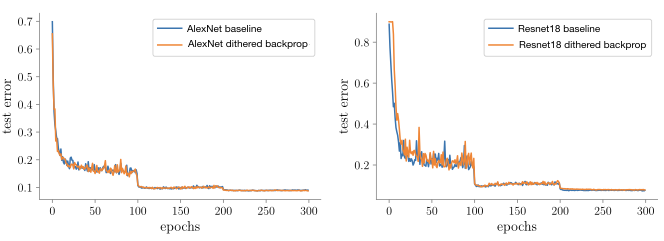}
	\caption{Test error of AlexNet and Resnet18 trained on CIFAR10 over the training epochs for baseline and dithered backpropagation.}
	\label{fig:more_acc}
\end{figure}

\begin{figure}
\centering
	\includegraphics[width=0.8\columnwidth]{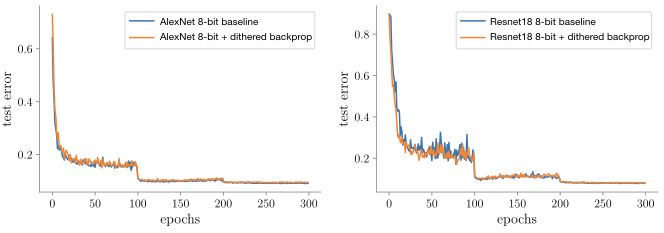}
	\caption{Test error of AlexNet and Resnet18 trained on CIFAR10 over the training epochs for 8-bit quantization and dithered 8-bit quantization.}
	\label{fig:more_acc2}
\end{figure}

Figures \ref{fig:more_acc} and \ref{fig:more_acc2} show the training curves of AlexNet and Resnet18 trained on CIFAR10 with the baseline method, dithered backprop, the reduced precision training method \cite{NIPS2018_7761}  and the combination of the latter two. As one can see, the training convergence is not affected by dithered backprop in any of the cases.

\subsubsection*{Comparison to meProp}
 \begin{figure}
	\centering
	\includegraphics[width=0.5\linewidth, height=50mm]{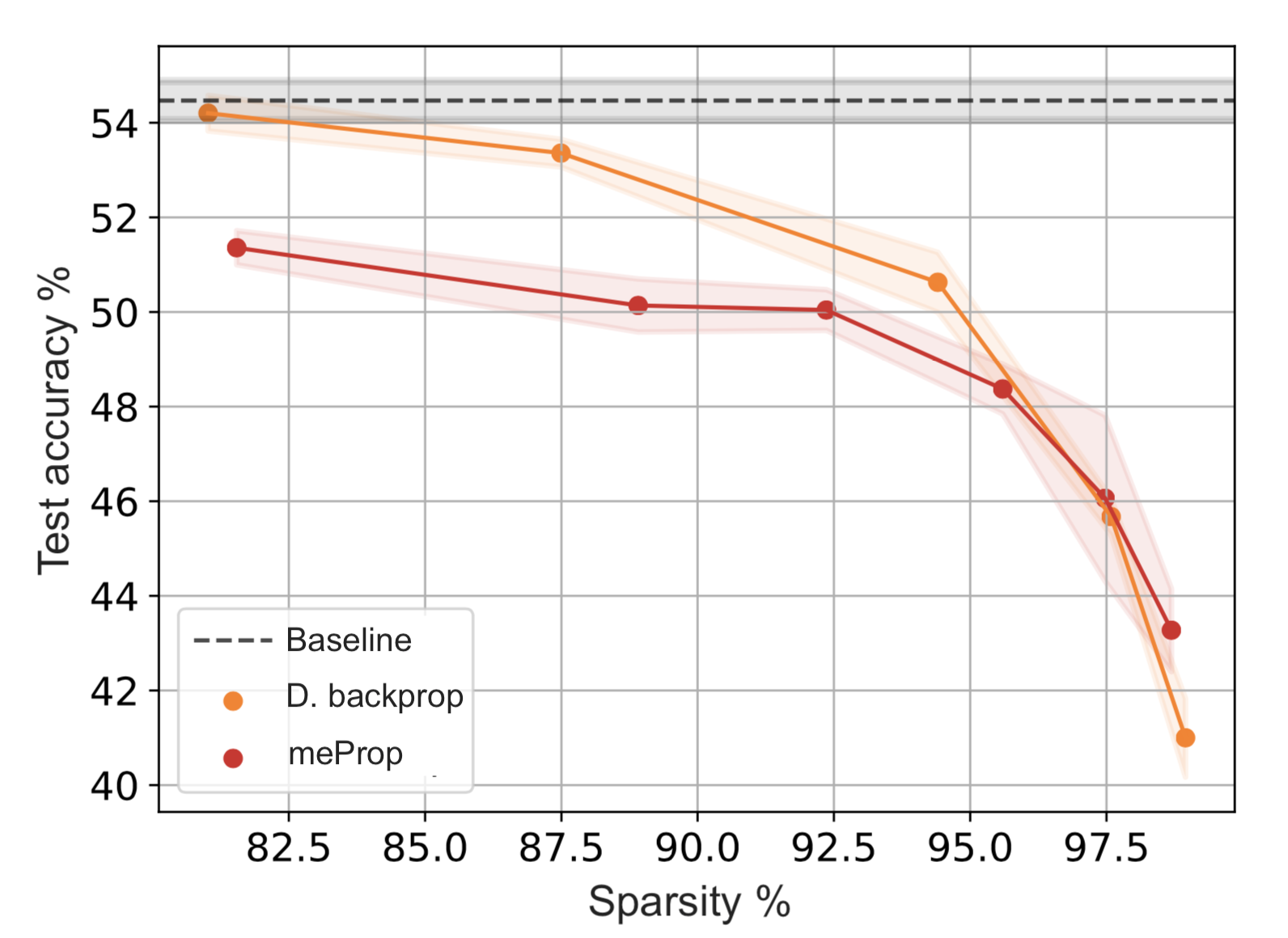}
	\caption{Learning performance at different levels of average sparsity of the preactivation gradients of a multilayer perceptron with two hidden layers (500, 500) trained on CIFAR10, using either regular back propagation (Baseline), dithered backprop (D. backprop) or meProp \cite{sun_meprop:_2017}. Multiple runs with different random seeds were executed for each configuration. Points show mean performance with standard deviation indicated as span.}
	\label{fig:cifar_mepropvsdb}
\end{figure}

In figure \ref{fig:cifar_mepropvsdb} we show the learning performance of the multilayer perceptron when trained on CIFAR10. As one can see, meProp does not reach as high accuracies as dithered backprop. We attribute this to the biased nature of their gradients estimates, which affects negatively the learning performance of the model.

\subsubsection*{Distributed training}
\begin{figure}
	\centering
	\includegraphics[width=0.5\linewidth, height=50mm]{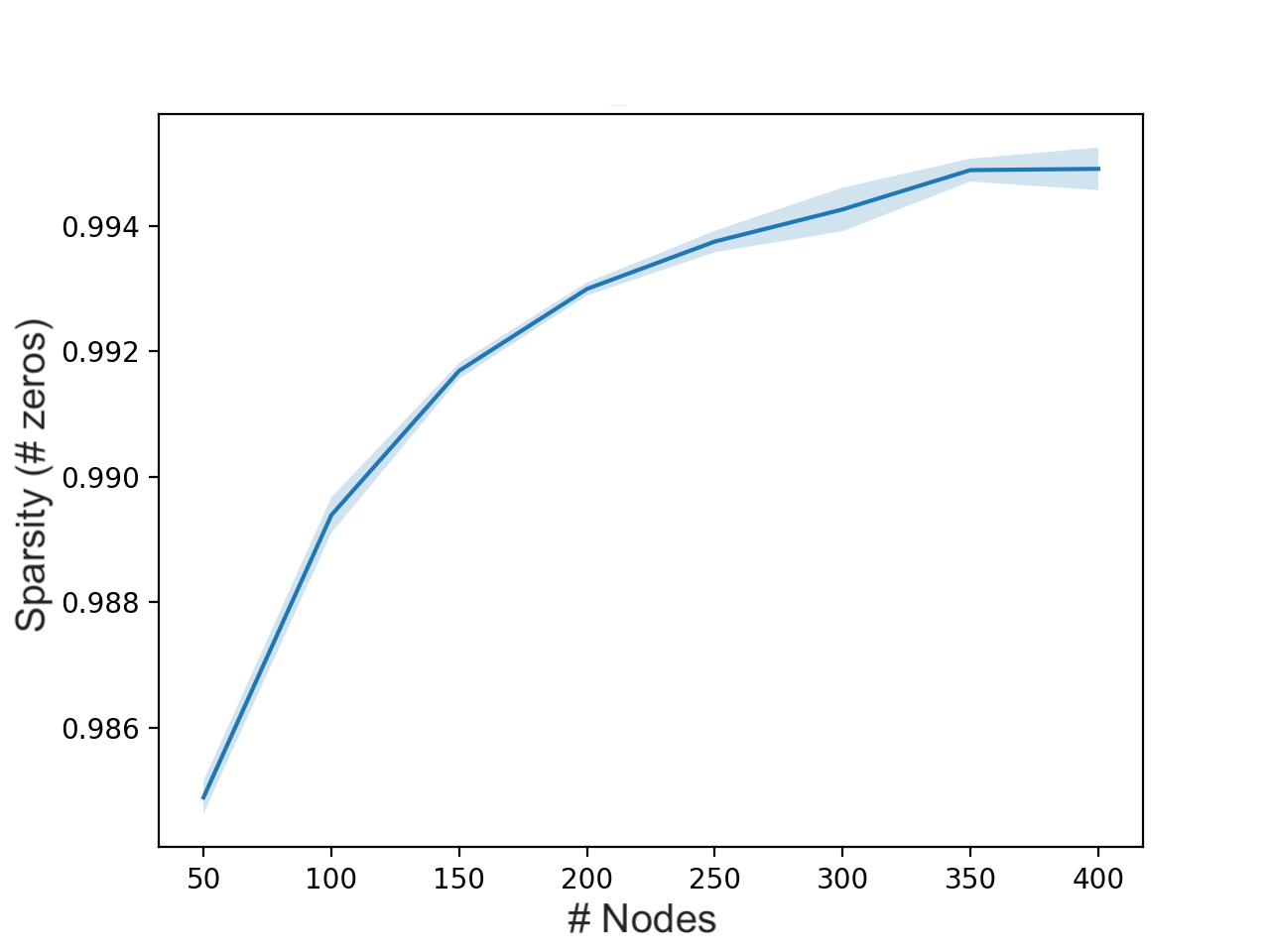}
	\caption{Average sparsity of the preactivation gradients of the convolutional layers of AlexNet trained on CIFAR10 with dithered backprop in a distributed training setting, at different number of participating nodes configuration. As the number of nodes increases, so does the sparsity at each node and therefore its computational savings for training. }
	\label{fig:conv_dt_sparse}
\end{figure}

\begin{figure}
	\centering
	\includegraphics[width=0.5\linewidth, height=50mm]{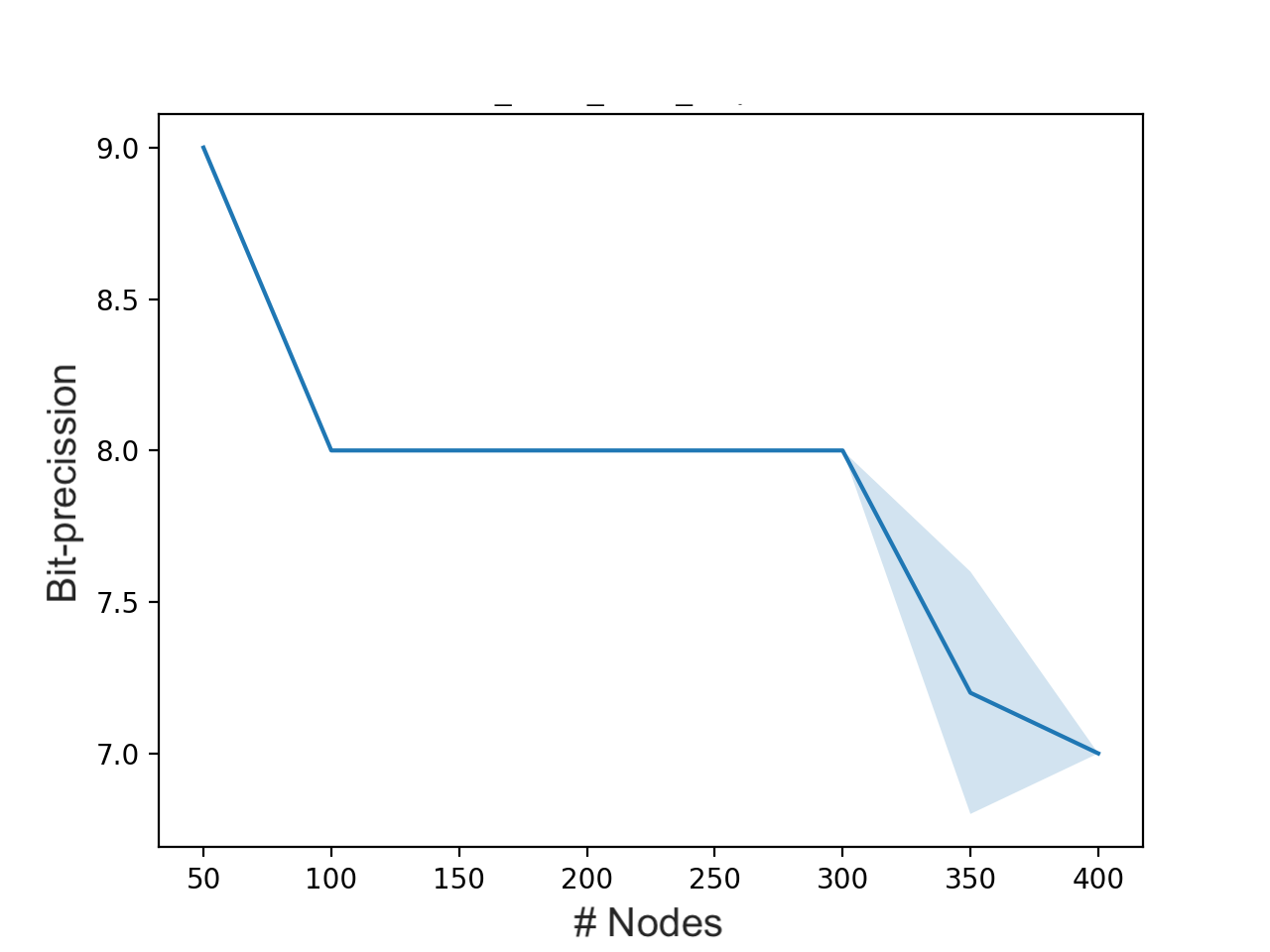}
	\caption{Maximal, worst-case bit-precision  of the convolutional layers of AlexNet trained on CIFAR10 with dithered backprop in a distributed training setting, at different number of participating nodes configuration. As the number of nodes increases, the number of bits necessary to represent the non-zero values decreases, and with it the computational cost for training at each node. }
		\label{fig:conv_dt_bits}
\end{figure}

Here we show the trend of the computational complexity of the convolutional layers as the number of participating nodes increases. As can be seen in figures \ref{fig:conv_dt_sparse} and \ref{fig:conv_dt_bits}, the computational decreases as the number of nodes increases.

\end{document}